%% file: main.tex
\pdfoutput=1

\documentclass[letterpaper, 10 pt, conference]{ieeeconf}  

\IEEEoverridecommandlockouts                              

\usepackage{graphicx} 
\usepackage{times} 
\usepackage{amsmath} 
\usepackage{amssymb}  
\usepackage{cite}
\usepackage{amsmath,amssymb,amsfonts}
\usepackage{algorithmic}
\usepackage{textcomp}
\usepackage{xcolor}
\usepackage{booktabs}
\usepackage{xspace}
\usepackage{multirow}
\usepackage{url}
\usepackage{tabularx}
\usepackage{makecell}
\let\labelindent\relax
\usepackage[inline]{enumitem}
\input{include/commands}

\title{\LARGE \bf
CoPlanVLM: Coordinated Path Planning for a Robot Team using Vision Language Models
}

\author{David R. Martin$^{1}$, Jared Paquet$^{2}$, Mohammad Abdullah Al Faruque$^{1}$, Pramod Khargonekar$^{1}$, \\
Kyle Volle$^{3}$, J. Humberto Ramos$^{2}$, and Scott Nivison$^{4}$
\thanks{$^{1}$ David Martin, Mohammad Abdullah Al Faruque, and Pramod Khargonekar are with the Department of Electrical Engineering and Computer Science, University of California, Irvine. }%
\thanks{$^{2}$ Jared Paquet and J. Humberto Ramos are with the Department of Mechanical and Aerospace Engineering at the University of Florida.}%
\thanks{$^{3}$ Kyle Volle is with Torch Technologies, LLC in Shalimar, FL.}%
\thanks{$^{4}$ Scott Nivison is with the Air Force Research Laboratory, Air Warfare Directorate, Eglin AFB Okaloosa County, Fl.}%
\thanks{This work was supported in part by the Air Force Research Laboratory, Air Warfare Directorate, Eglin AFB, FL.}
}

\begin{document}

\maketitle
\thispagestyle{empty}
\pagestyle{empty}


\begin{figure*}[t]
    \centering
    \includegraphics[width=\linewidth]{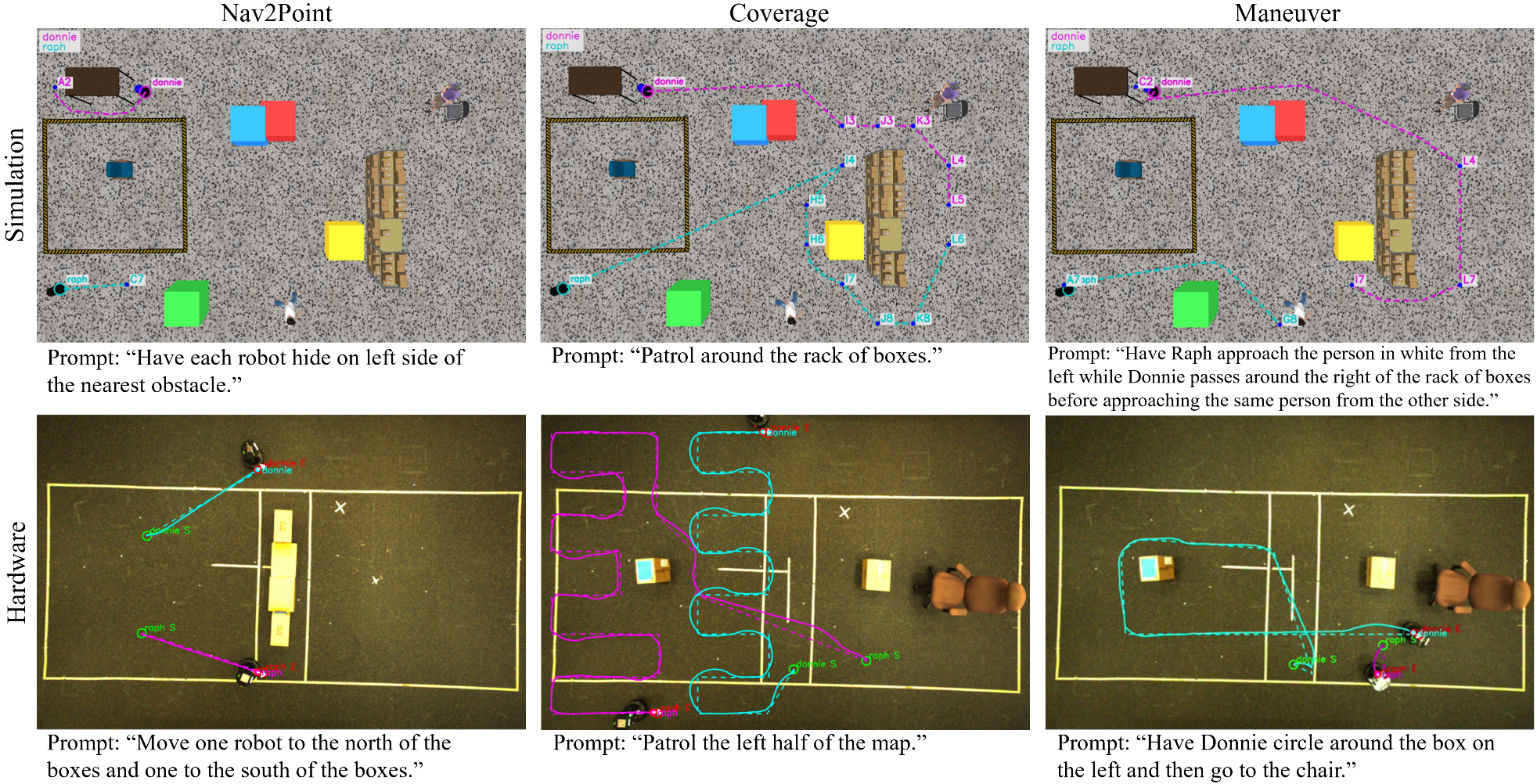}
    \caption{Paths planned by \ourmethod on six different tasks in simulation and hardware  experiments. Two tasks are shown from each of the task categories \textit{Nav2Point}, \textit{Coverage}, and \textit{Maneuver}. \textbf{(Top)} Offline testing results on three prompts in the Gazebo environment. Dotted lines indicate planned paths and blue dots with alphanumeric labels mark the waypoints selected by the VLM. \textbf{(Bottom)} Hardware test results on three prompts. Dotted lines indicate planned paths, and solid lines mark the true path followed by the robots with magenta and blue paths corresponding to Raph and Donnie respectively. The robots' start locations are labeled in green, with stop locations in red.}
    \label{fig:full_results_image}
\end{figure*}

\begin{abstract}
    We present \ourmethod, which uses a Vision Language Model (VLM) to translate a single natural-language instruction and image into coordinated paths for a team of two robots. While VLMs offer a flexible, natural-language interface for commanding robot teams, their weak spatial reasoning has greatly limited VLM-based path planning. \ourmethod improves the VLM's spatial reasoning and path planning abilities through three components: (1) an image overlay marking both free space and obstacles, (2) task classification that selects the most appropriate low-level planner and reasoning prompt for each instruction, and (3) task-specific chain-of-thought reasoning that prompts the VLM to identify key objects, allocate roles, and verify routes are collision-free before committing to a plan. Together, these components enable the team to complete complex coordinated maneuvers requiring both robots and divide large tasks between robots to increase efficiency. \ourmethod also supports detailed maneuvers involving specific routes, loops, and coverage of large areas, as well as tasks requiring directional reasoning such as north/south or left/right. We validate \ourmethod in both Gazebo simulation and on physical Turtlebot 4 hardware. \ourmethod attains an average success rate of 84.4\% compared to 50.0\% for the closest baseline. On trials where both methods succeed, \ourmethod's paths are 27\% shorter on average. Project code is available at \textit{github.com/AICPS/CoPlanVLM}.
\end{abstract}

\section{Introduction}

Multi-robot teams have shown great promise in diverse application domains such as warehousing \cite{agaskar2025deepfleet}, UAV swarms \cite{rahman2025multiuavsurvey}, and household tasks \cite{liu2025coherent}. Through collaboration, teams are able to complete complex missions which may be impossible for individuals \cite{liu2025coherent, borate2025CoMuRoS, karam2026collaboration}. However, planning for a collaborative team presents a significant challenge that requires understanding mission intent, allocating roles, and planning each robot's motion. Large Language Models (LLMs) have been proposed to address these challenges by enabling robotic systems to interpret complex human instructions, coordinate multiple robots, and adapt to diverse contexts~\cite{li2026survey}. Vision Language Models (VLMs) can additionally use camera images from the robots or workspace as context to inform planning. LLMs and VLMs have been proposed to generate high-level team plans~\cite{borate2025CoMuRoS, liu2025coherent}, and navigate individual robots directly from natural-language instructions ~\cite{sathyamoorthy2024convoi, potnis2026catnav, bao2026goal2pixel}.

However, VLMs often exhibit poor spatial reasoning~\cite{stogiannidis2025mind}, which limits their ability to generate long-horizon, collision-free paths. Consequently, many approaches do not construct paths with the VLM directly, instead using it to identify goal locations and/or regions to avoid, which form a cost map over which a classical planner routes to the target~\cite{elnoor2025gronav, rana2026overseec, potnis2026catnav}. While this reduces reliance on VLM spatial reasoning, it reduces the problem to reaching a goal at lowest cost, and struggles to express more complex behaviors. Many approaches to Vision Language Navigation repeatedly query a VLM for the next action~\cite{goetting2024end, tian2024drivevlm, sathyamoorthy2024convoi, bao2026goal2pixel}, or use Vision-Language-Action models trained for next-action prediction~\cite{wang2025trackvla, cheng2024navila}. However, these methods do not produce full long-horizon plans and require frequent querying of large models. The authors of \cite{sathyamoorthy2024convoi} and \cite{pasqualina2026vision} plan paths by prompting a VLM to select ordered waypoints, but both target single robots, with \cite{sathyamoorthy2024convoi} navigating only to a predefined destination and \cite{pasqualina2026vision} providing no mechanism to ensure obstacle-free paths. To our knowledge, no prior work generates long-horizon paths for a collaborative team directly from natural-language instructions and supports complex maneuvers beyond point-to-point navigation.

\textbf{To address this gap, this work introduces \ourmethod, which leverages a commercially available VLM to plan coordinated maneuvers for a collaborative team of two robots following a single natural-language instruction.} It is tested on tasks explicitly requiring both robots (e.g. approaching the same target from two different directions), tasks where coordination improves efficiency (e.g. dividing a region to patrol efficiently), and tasks requiring no coordination. Additionally, \ourmethod enables the performance of detailed maneuvers involving specific routes, loops, and surveying large areas. Experimental results in both simulation and real-world scenarios demonstrate that \ourmethod attains an average success rate of 84.4\% compared to 50.0\% for the closest baseline \cite{sathyamoorthy2024convoi}, and on trials where both methods succeed, \ourmethod's paths are 27\% shorter on average. Hardware experiments demonstrate successful deployment on two Turtlebot 4 robots. \ourmethod improves the spatial reasoning and path planning capabilities of the VLM through:

\begin{enumerate}
    \item Image preprocessing that marks free space and detected objects on the image, improving the VLM's spatial grounding and steering it away from obstacles;
    \item Task classification that adaptively selects the best reasoning prompt and low-level path planner for each instruction, spanning point-to-point navigation, area coverage, and route-constrained maneuvers;
    \item Task-specific chain-of-thought reasoning in which the VLM identifies key objects, allocates roles, and verifies that selected waypoints are unobstructed.
\end{enumerate}

\section{Related Work}

\subsection{LLM Task Planning For Robot Teams}

Large Language Models (LLMs) and Vision-Language Models (VLMs) are increasingly proposed for collaborative robotic systems due to their ability to interpret natural-language instructions and generalize across a wide variety of tasks~\cite{li2026survey}. A common paradigm employs a centralized planner LLM that decomposes a mission and assigns natural-language subgoals to individual robots, each equipped with a local LLM that reasons about how to accomplish its assigned subtask~\cite{borate2025CoMuRoS, liu2025coherent, safety_aware_planning}. However, coordinated path planning using language models remains largely underexplored. The authors in \cite{yu2023conavgpt} and \cite{wang2025comres} employ a VLM to coordinate a robot team's exploration of an unknown region in search of a target object, but the VLM only selects the most promising frontier to explore next, limiting the system's ability to generalize to other tasks. The authors in \cite{pasqualina2026vision} use a VLM to plan a path for a single robot, but the proposed method does little to ensure path feasibility and does not coordinate its actions with any other robots. \ourmethod expands the generalizability of language models to long-horizon coordinated path planning for a team of two robots.

\subsection{Vision Language Navigation}

Vision-Language Navigation (VLN) is the problem of converting a free-form natural-language instruction into a sequence of movement decisions that produce a matching path~\cite{anderson2018vln}. One common approach repeatedly queries a VLM for the next action the robot should take~\cite{goetting2024end, tian2024drivevlm, sathyamoorthy2024convoi, bao2026goal2pixel}, and more recently, Vision-Language-Action (VLA) models have been trained to output navigation commands directly~\cite{wang2025trackvla, cheng2024navila}. However, both strategies produce only the immediate next action rather than a long-horizon plan, which hinders coordination between robots and limits their ability to execute long, complex maneuvers.

Another common approach uses a VLM to identify target locations and regions to avoid, producing a cost map over which a classical planner computes a path~\cite{elnoor2025gronav, rana2026overseec, potnis2026catnav}. This yields efficient, long-horizon paths, but a cost-map planner always returns the shortest, lowest-cost route to the goal, which limits the system to instructions whose solution is simply the most direct path. It cannot express behaviors where the route itself matters, such as patrolling a region, looping around an object, or passing on a specified side of an obstacle (e.g., staying to the right of a rack of boxes). In this work, the VLM instead builds a complete route by selecting as many intermediate waypoints as the task requires, enabling these more complex behaviors.

\begin{figure*}[htb]
    \centering
    \includegraphics[width=\linewidth]{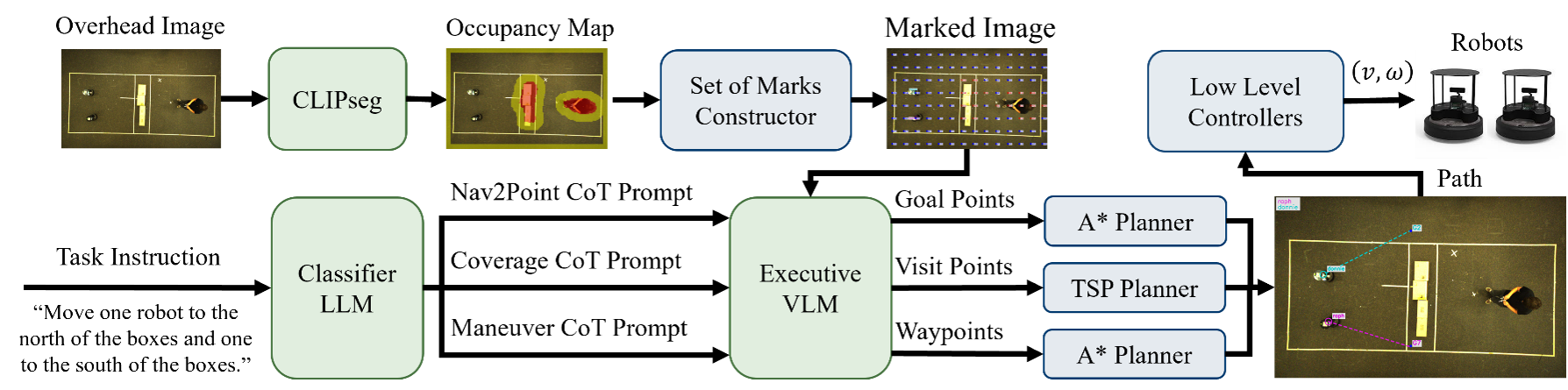}
    \caption{\ourmethod methodology overview. First an overhead image is processed by CLIPSeg \cite{luddecke2022image} to detect obstacles and create the occupancy map. A classifier LLM classifies each instruction into one of three categories (Nav2Point, Coverage, Maneuver)  which determines the corresponding executive VLM prompt and low-level path planning algorithm. The executive VLM selects marked regions on the image which are used as inputs to the low-level path planning algorithms. These paths are then processed by low level controllers to send commands to the robots.}
    \label{fig:methods}
\end{figure*}

\subsection{Spatial Reasoning for VLMs} 

A key challenge of using VLMs for path planning is their poor spatial reasoning~\cite{stogiannidis2025mind}, which often leads them to choose paths that are inconsistent with the instruction, inefficient, or that pass directly through obstacles. Set-of-Mark prompting improves spatial reasoning by overlaying numbered markers on an image ~\cite{yang2023setofmark}. This technique has been applied to path planning in several ways: \cite{pasqualina2026vision} overlays a labeled grid from which the VLM selects regions, while \cite{sathyamoorthy2024convoi} places numbered marks only in free space. Similarly, \cite{bai2025room2chess} converts an agent's environment into a grid-based chessboard to improve an LLM's spatial reasoning for low-level control. Building on these ideas, we employ a unique marking approach using a similar chessboard-style notation with distinct markings for free space and objects of interest.

\section{Problem Formulation}
\ourmethod plans coordinated paths for a team of two Turtlebot 4\footnote{\url{https://robots.ros.org/turtlebot4/}} robots, named Raph and Donnie, operating in a shared workspace. Let $\mathcal{R} = \{\text{Raph}, \text{Donnie}\}$ denote the robot team.

\textbf{Inputs:} At planning time, the system is given:
\begin{enumerate}[leftmargin=*, itemsep=0pt, topsep=2pt]
    \item an overhead RGB image $I$ of the workspace;
    \item the pose of each robot $r \in \mathcal{R}$, denoted $q_r = (x_r, y_r, \theta_r)$;
    \item a natural-language instruction $L$ describing the task.
\end{enumerate}

The overhead camera image $I$ provides a top-down view of the environment which is common in robot planning. Such views arise from overhead cameras in structured settings \cite{pasqualina2026vision}, aircraft or satellite imagery outdoors~\cite{elnoor2025gronav}, from generated top-view maps~\cite{yu2023conavgpt} and 2D occupancy maps~\cite{wang2025comres}.

\textbf{Outputs:}
\ourmethod leverages a VLM and low-level planners to produce an ordered sequence of waypoints,
\begin{equation}
    W_r = \left( w_r^{1}, w_r^{2}, \dots, w_r^{n_r} \right), \qquad w_r^{k} = (x_r^{k}, y_r^{k}) \in \mathcal{W},
\end{equation}
for each robot $r \in \mathcal{R}$, where $n_r$ is the number of waypoints assigned to robot $r$. Executing $W_r$ drives robot $r$ sequentially through its waypoints. We denote the joint plan by $W = \{W_{\text{Raph}}, W_{\text{Donnie}}\}$.

\textbf{Objective:}
The goal is to produce a joint plan $W$ that accomplishes the instruction $L$ while ensuring the straight-line segment between each pair of consecutive waypoints avoids all obstacles. Planned paths are permitted to intersect as inter-robot collision avoidance is enforced after planning through a control barrier function. Among obstacle-free plans that accomplish $L$, we seek the one that minimizes the makespan, estimated as the longer of the two robots' total path lengths $\mathrm{MS}(W) = \max\left( \lVert W_{\text{Raph}} \rVert, \; \lVert W_{\text{Donnie}} \rVert \right)$, where $\lVert W_r \rVert$ is the total length of robot $r$'s waypoint path.

\section{Methods}

The overall system architecture is depicted in Figure~\ref{fig:methods}. The overhead image is first preprocessed with CLIPSeg~\cite{luddecke2022image} to produce an occupancy map and a marked image for prompting the VLM. A classifier LLM then categorizes the instruction and selects the corresponding VLM prompt and low-level planner. The VLM then selects key waypoints from the marked image for each robot which are combined into efficient, collision-free paths through the low-level planners.

\subsection{Image Pre-processing}
The overhead image is preprocessed by CLIPSeg \cite{luddecke2022image} to detect obstacles  and construct both an occupancy map and a marked image to prompt the VLM.

\subsubsection{Semantic Segmentation}

CLIPSeg~\cite{luddecke2022image} is a vision-language model that uses CLIP as a backbone to perform prompt-based image segmentation. Given an image and a text prompt $p$, CLIPSeg produces a dense segmentation score $C(q, p)$ for each pixel $q$, indicating the degree to which that pixel corresponds to the concept described by $p$. \ourmethod is provided a set of traversable prompts $P_{\text{trav}}$ describing navigable terrain, and any pixel whose confidence falls below $0.50$ for all such prompts is conservatively treated as untraversable. In this work, all testing is in indoor environments, motivating the use of a single traversable prompt, $P_{\text{trav}} = \{\text{``the floor''}\}$. Any pixel with confidence at least $0.50$ for ``the floor'' is treated as traversable and all others as untraversable.

\subsubsection{Occupancy Map Construction}\label{sec:occupancy_map}
\begin{figure}[h]
    \centering
    \includegraphics[width=.90\linewidth]{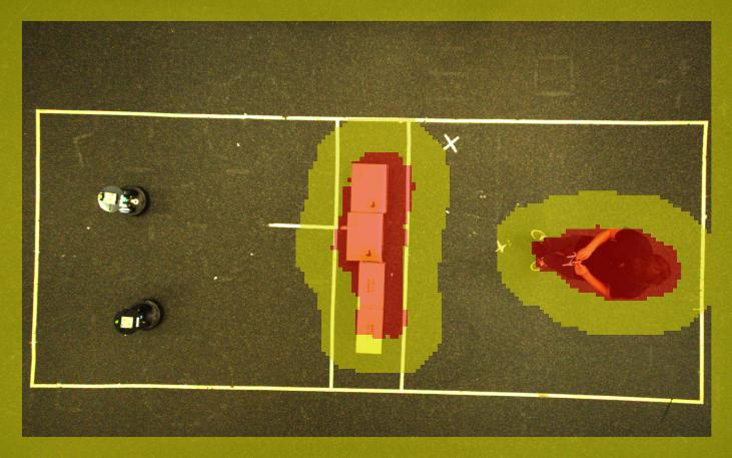}\\[2pt]
    \includegraphics[width=.90\linewidth]{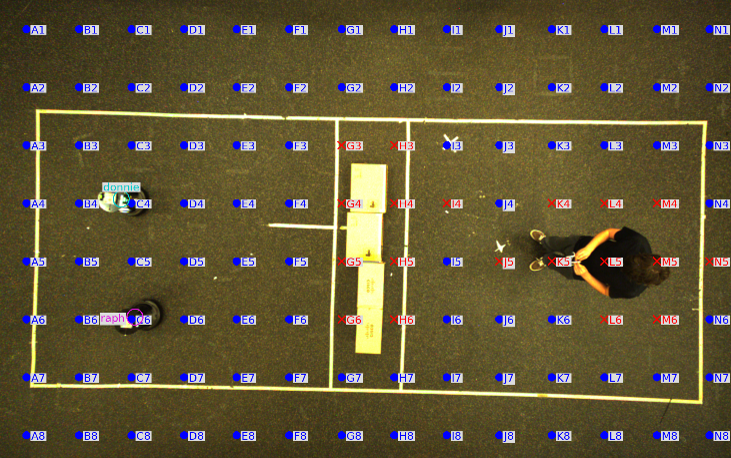}
    \caption{Image preprocessing. \textbf{(Top)} Occupancy map visualization with obstacles detected by CLIPSeg shown in red, and obstacle inflation and the map-edge buffer in yellow. \textbf{(Bottom)} Marked image passed to the executive VLM, with free space marked by blue dots, obstacles by red ``x'' marks, and robots labeled with their names (``raph'' and ``donnie'').}
    \label{fig:image_processing}
\end{figure}

To construct an occupancy grid for planning, each CLIPSeg pixel coordinate $(p_x, p_y)$ is converted to a world position $(x, y)$ using a pinhole model of the overhead camera together with its intrinsics and extrinsics. Assuming a flat floor at constant height $h$ below the camera, the projection is
\begin{equation}
    x = \frac{(p_x - c_x)\, h}{f_x}, \qquad
    y = -\frac{(p_y - c_y)\, h}{f_y},
\end{equation}
where $(f_x, f_y)$ are the camera's focal lengths, $(c_x, c_y)$ is the principal point, and $h$ is the height of the camera above the floor. The projected obstacle locations form an occupancy map at $0.05$~m resolution (Figure~\ref{fig:image_processing}, Top), to which several corrections are applied. Because CLIPSeg is unreliable near the image boundary, detections within $65$ pixels of the edge are discarded, obstacles are inflated by $0.45$~m for the robot footprint, and cells within $0.25$~m of a map edge are marked untraversable. To prevent robots from being detected as obstacles, any detections within a $0.65$~m radius ($0.2$~m robot radius + $0.45$~m obstacle inflation radius) of each robot's pose are cleared before inflating the obstacles. 

\vspace{-5pt}
\subsubsection{Mark Overlay Construction} \label{sec:setofmarks}
A mark overlay (Figure~\ref{fig:image_processing}, Bottom) is constructed by superimposing an $8 \times 14$ grid of labeled points onto the image. Each point is assigned a chessboard-style alphanumeric label (e.g., ``C5''; letter for column, number for row) that encodes its grid position symbolically, allowing spatial relationships to be expressed through text. To remove directional ambiguity, the prompt fixes a mapping between cardinal directions and image axes, associating north, south, east, and west with up, down, right, and left, respectively. Free points are rendered as blue dots and blocked regions as red ``x'' marks, with the executive VLM prompted to select navigation waypoints from the free points. Finally, each robot is labeled with its name at the image location corresponding to its current pose, enabling the VLM to distinguish between them.

\subsection{Task Classification} \label{sec:task_classification}
Each task instruction is first categorized by a classifier LLM into one of three categories. The descriptions provided to the LLM are as follows:

\textit{Nav2Point}: ``Best for tasks where each robot drives to one or more goal locations in order. The destination(s) matter, but the exact path taken does not.''

\textit{Coverage}: ``Best for tasks that involve patrolling or sweeping an area. The planner selects high-level regions to cover rather than a specific path, and a low-level controller ensures all selected regions are visited efficiently.''

\textit{Maneuver}: ``Best for tasks where the specific route matters, not just the destination. Used when the instruction constrains \emph{how} the robot travels, such as loops, specified routes, or avoidance, ...etc. This category also functions as a catch-all category for tasks that do not fall neatly into the \textit{Nav2Point} or \textit{Coverage} categories.''

\subsection{Executive VLM Prompting}

A unique prompt is designed for each of the three task categories listed in Section~\ref{sec:task_classification} and is sent to the executive VLM which selects waypoints from the marks on the overhead image. Full prompts for each of the three classes are available in the project's GitHub repository. Each prompt sent to the VLM is assembled from six components, in order:

\begin{enumerate}[leftmargin=*, itemsep=0pt, topsep=2pt]
    \item \textbf{Preamble} -- robot descriptions, image context, and global rules shared across all tasks.
    \item \textbf{Map Overlay Description} -- an explanation of what the markings on the overlaid image represent.
    \item \textbf{Controller Explanation} -- a per-category block describing how the controller works, how to select points, the output schema, and a worked example.
    \item \textbf{Operator Instruction} -- Natural-language instruction $L$.
    \item \textbf{Chain-of-Thought Scaffold} -- a per-category reasoning scaffold guiding the VLM's reasoning.
    \item \textbf{Marked Image} -- the marked overhead image.
\end{enumerate}

\subsection{Chain of Thought Scaffold}

Before selecting a route, the executive VLM outputs a reasoning trace in text. The CoT reasoning scaffolds differ between the task categories, but follow a similar structure:
\begin{enumerate}[leftmargin=*, itemsep=0pt, topsep=2pt]
    \item \textbf{Grounding} - Locates each robot, identifies key objects in the instruction, labels the object at each red ``x'' marker. 
    \item \textbf{Task Allocation} - Divides the task between the robots by  identifying which robot is best suited for each part.
    \item \textbf{Leg Decomposition} (Maneuver only) -- Splits each robot's assigned task into an ordered sequence of legs (e.g., travel to chair, loop around it, travel to person).
    \item \textbf{Route Construction} - Lists waypoints for each robot.
    \item \textbf{Verify} - Verifies no selected point is blocked, replacing any that are with a nearby unblocked point.
\end{enumerate}

\subsection{Low-Level Planners}
Low-level path planners are used to convert the VLM chosen waypoints into efficient paths that avoid blocked regions in the occupancy map (Section~\ref{sec:occupancy_map}).

\subsubsection{A* Path Planning}\label{sec:astar_planning}
For \textit{Nav2Point} and \textit{Maneuver} tasks, the executive VLM outputs an ordered list of waypoints for each robot: a single goal or a few goals for \textit{Nav2Point}, and a longer sequence specifying a precise route for \textit{Maneuver}. The A*~\cite{astar} algorithm then plans collision-free paths through these waypoints in order, beginning from each robot's initial position. The image marking and chain-of-thought reasoning are designed to steer the VLM away from blocked regions, but in the case a blocked waypoint is selected, the planner first attempts to project that waypoint to the nearest free cell within a $1$m radius before dropping it if none exists. This ensures safety while attempting to preserve VLM intent although poor VLM selections can still lead to task failures.

\subsubsection{Coverage Path Planning}
For \textit{Coverage} tasks, the robot's sensing footprint is assumed to match the marking resolution of the overlay, so every marked point within the specified region must be visited, though the visiting order is unconstrained. The VLM outputs an unordered set of regions for each robot, and the planner computes an efficient visiting order by solving an open-path traveling salesman problem anchored at the robot's initial pose, using nearest-neighbor construction~\cite{rosenkrantz1977analysis} refined by 2-opt~\cite{croes1958method}. Edge costs are shortest-path distances on the inflated occupancy grid, computed with Dijkstra's algorithm~\cite{dijkstra2022note}.

\begin{table*}[hbt]
\setlength{\tabcolsep}{2pt}
\centering
\caption{\ourmethod success rates and average makespan for all evaluation prompts (10 trials per prompt).}
\label{tab:full_prompt_results}
\renewcommand{\arraystretch}{1.1}
\begin{tabularx}{\textwidth}{c c c c c X}
\toprule
\textbf{Prompt} & \textbf{Category} & \textbf{Env.} & \textbf{SR (\%)} & \makecell{\textbf{MS} \textbf{(m)}} & \textbf{Prompt} \\
\midrule
1  & Nav2Point & Sim & 100.0 & 14.6 & ``Surround the person in the purple shirt by sending robots to the left and right sides of the person.'' \\
2  & Nav2Point & Sim & 80.0  & 3.6  & ``Have each robot hide on the left side of the nearest obstacle.'' \\
3  & Nav2Point & Sim & 80.0  & 8.0  & ``Visit each of the four colored boxes.'' \\
4 & Nav2Point & AVL & 100.0 & 3.7  & ``Send both robots to the white X marked in tape.'' \\
5 & Nav2Point & AVL & 100.0 & 2.9  & ``Move one robot to the north of the boxes and one to the south of the boxes.'' \\
6 & Nav2Point & AVL & 50.0  & 2.8  & ``Visit each object in the scene.'' \\
7  & Coverage & Sim & 90.0  & 34.4 & ``Patrol the perimeter of the map.'' \\
8  & Coverage & Sim & 90.0  & 17.6 & ``Patrol around the rack of boxes.'' \\
9  & Coverage & Sim & 70.0  & 9.8  & ``Survey the area inside the black and yellow striped tape.'' \\
10 & Coverage & AVL & 90.0  & 5.9  & ``Have the robots patrol the area around the chair.'' \\
11 & Coverage & AVL & 70.0  & 11.5 & ``Have the robots survey the large square marked in tape on the left side.'' \\
12 & Coverage & AVL & 70.0  & 22.7 & ``Patrol the left half of the map.'' \\
13  & Maneuver & Sim & 100.0 & 19.2 & ``Have Donnie do a loop around the red and blue boxes and then meet up with Raph.'' \\
14  & Maneuver & Sim & 80.0  & 18.7 & ``Have Raph approach the person in white from the left while Donnie passes around the right of the rack of boxes before approaching the same person from the other side.'' \\
15  & Maneuver & Sim & 100.0 & 24.0 & ``Have one robot move along the north edge of the map to the person in purple while the other robot travels along the south edge until it is directly south of the same person, then move north to the person.'' \\
16 & Maneuver & AVL & 100.0 & 6.8  & ``Have the robots switch places. Have Raph move around the south of the chair and have Donnie move around the north of the chair.'' \\
17 & Maneuver & AVL & 100.0 & 7.5  & ``Send both robots to the person. Have one approach around the north of the boxes, and one approach around the south of the boxes.'' \\
18 & Maneuver & AVL & 50.0  & 11.7 & ``Have Donnie circle around the box on the left and then go to the chair.'' \\
\bottomrule
\end{tabularx}
\end{table*}

\begin{figure*}[htb]
    \centering
    \includegraphics[width=\linewidth]{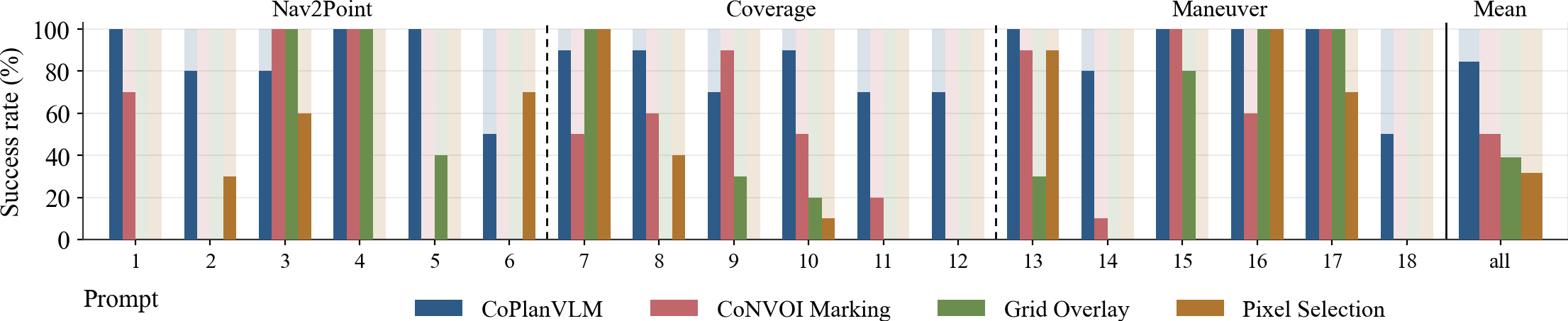}
    \caption{Measured success rates for \ourmethod and compared methods on all 18 prompts.}
    \label{fig:sucess_rates_histogram}
\end{figure*}

\subsection{Low-Level Control}

\subsubsection{Path Tracking}
Each robot follows its planned path with a $10$~Hz proportional controller that steers toward the active goal and sets forward speed proportional to the remaining distance. Forward motion is gated on heading error: while the goal lies more than $\psi_{\max}=50^\circ$ off the robot's heading, the robot turns in place before driving, which avoids wide, corner-cutting arcs. Goals are advanced within a $0.1$~m capture radius.

\subsubsection{Safety filter.}
To avoid collisions between robots, the nominal command $u_{\text{nom}} = [v, \omega]^\top$ is filtered by a control-barrier-function (CBF) quadratic program. Following \cite{wilson2020robotarium}, we enforce the barriers on a look-ahead point $\ell = 0.10$~m ahead of the robot center. We impose two barriers: one on the peer robot and one on the nearest occupied cell of the occupancy map, ensuring the robot does not drift into obstacles while avoiding its teammate. For each hazard $j \in \{\text{peer}, \text{obstacle}\}$, let $\Delta d_j$ be the vector from the look-ahead point to the hazard and define the barrier $h_j = \lVert \Delta d_j \rVert^2 - R_j^2$, which is positive when the look-ahead point lies outside the safety radius $R_j$. Each barrier is enforced through the CBF condition $\dot{h}_j(x, u) \geq -\gamma\, h_j$, and the filtered command is the solution of the quadratic program:
\begin{equation}
    \begin{aligned}
        u^\star = \arg\min_{u} \;\; & \lVert u - u_{\text{nom}} \rVert^2 \\
        \text{s.t.} \;\; & \dot{h}_{\text{peer}}(x, u) \geq -\gamma\, h_{\text{peer}}, \\
        & \dot{h}_{\text{obs}}(x, u) \geq -\gamma\, h_{\text{obs}}.
    \end{aligned}
\end{equation}

\section{Experiments}\label{sec:experimental_setup}

\ourmethod is evaluated on two Turtlebot 4 robots in two settings: a simulated warehouse in Ignition Gazebo Fortress~\cite{koenig2004design} populated with people, furniture, and boxes (Figure~\ref{fig:full_results_image}~Top), and real-world experiments with two physical Turtlebots at the University of Florida Autonomous Vehicle Laboratory (AVL) (Figure~\ref{fig:full_results_image}~Bottom). GPT-4o (OpenAI) served as both the classifier LLM and executive VLM in \ourmethod, and as the VLM for all baselines.

\subsection{Tasks}\label{sec:prompts}
\ourmethod was evaluated on 18 prompts: six \textit{Nav2Point} tasks (travel to specified points), six \textit{Coverage} tasks (patrol or survey a region), and six \textit{Maneuver} tasks (follow a specified route), all listed in Table~\ref{tab:full_prompt_results} with half of each task category evaluated in simulation and half in the AVL. The tasks were designed to test different kinds of coordination. Some require both robots to complete the task (prompts 1, 4, 5, 13--17), some can be made more efficient by dividing the work between the robots (prompts 3, 6, 7--12), and some require little coordination (prompts 2, 18). Several tasks (prompts 1, 2, 5, 12, 14--18) also require spatial reasoning about direction, such as north/south or left/right.

\subsection{Offline Testing}\label{sec:offline_testing}
Because large-scale evaluation on physical robots is time-intensive, the comparison against baseline methods and the ablation study were conducted offline. In this setting, each method is prompted with an overhead image of the workspace and produces a planned path for the robots. Without executing the plan on the robots, each path is evaluated directly for whether it completes the task instruction and avoids obstacles. Each of the 18 prompts was run ten times per method with results reported in Table~\ref{tab:full_prompt_results} and Figure~\ref{fig:sucess_rates_histogram}.

\subsection{Performance Metrics}
All methods were evaluated on two metrics: success rate (SR), which measures whether the task was completed, and normalized makespan, which estimates completion time.

\textbf{Success Rate} evaluates whether the planned route completes the task described in the instruction without intersecting any obstacles. Success criteria depend on the task, with evaluation criteria for each task provided on the project GitHub. For \textit{Coverage} tasks, success additionally requires at least 95\% coverage of the grid points in the described area.

\textbf{Normalized Makespan} measures the efficiency of the planned paths from their lengths. Because the two robots execute their paths in parallel, the makespan of a single trial is the longer of the two path lengths, $\mathrm{MS} = \max(\ell_{\text{Donnie}},\, \ell_{\text{Raph}})$. Different prompts demand very different amounts of travel, so raw makespans are not directly comparable across tasks. We therefore define the Normalized Makespan $NMS(M,L)$ of method $M$ on task $L$ as its mean makespan divided by that of \ourmethod on the same task,
\begin{equation}
    \mathrm{NMS}(L, M) = \frac{\overline{\mathrm{MS}}(L, M)}{\overline{\mathrm{MS}}(L, \ourmethod)},
\end{equation}
where $\overline{\mathrm{MS}}(L, M)$ is the mean makespan of method $M$ on task $L$. Only successful trials are included in each average. If a baseline fails every trial for a given task, that task is excluded from its average. Consequently, each baseline is compared to \ourmethod on a different subset of prompts. By construction, \ourmethod has a normalized makespan of $1.0$, and values above (or below) $1.0$ indicate longer (or shorter) paths relative to \ourmethod.

\subsection{Comparison to Existing Methods}\label{sec:compared_methods}
Because no prior work has investigated coordinated path planning from natural language for the types of complex behavior investigated in this work, \ourmethod was compared against three representative approaches for VLM-based path planning that were adapted to the two-robot setting.

\textbf{Grid Overlay:} The closest work to our problem setting is~\cite{pasqualina2026vision}, which uses a VLM to plan paths for a single robot executing complex behaviors described in natural language. A labeled grid is overlaid on the overhead image, and the VLM selects an ordered list of cells whose centroids serve as a sequence of target locations for a PID controller. We adapt this approach to our multi-robot setting by labeling each robot with its name and prompting the VLM to plan routes for both. Because \cite{pasqualina2026vision} relies on the VLM alone to select obstacle-free routes, initial testing showed very high collision rates. We therefore strengthen this baseline by enforcing obstacle avoidance with the A* planner of Section~\ref{sec:astar_planning} for a fairer comparison of the marking and prompting methods.
 
\textbf{CoNVOI Marking:} CoNVOI~\cite{sathyamoorthy2024convoi} performs single-robot navigation from an ego-centric viewpoint. It uses a VLM to identify the robot's context (e.g., indoor vs.\ outdoor) and select appropriate navigation rules. CoNVOI marks only obstacle-free regions of the image with numbers from which the VLM selects navigation waypoints. CoNVOI is adapted to our setting by fixing a single rule set across all prompts, since our scenes are all similar indoor environments. It is extended to two robots by labeling each with its name and prompting the VLM to select a list of waypoints for each to follow. The marking scheme is replicated identically by placing marks only in free cells in the occupancy map.

\textbf{Pixel Selection:} An alternative to selecting from a set of marks is to have the VLM select waypoints as pixel coordinates on an unmarked image as in ~\cite{bao2026goal2pixel, hu2025see}. Although these works target single robots, we adapt pixel-coordinate selection to our problem by prompting the VLM to output a sequence of pixel coordinates $(p_x, p_y)$ for each robot on a normalized $1000 \times 1000$ scale as in~\cite{hu2025see}. These are projected to world coordinates $(x, y)$ and used as waypoints in the same A* planner introduced in Section~\ref{sec:astar_planning}.

Table~\ref{tab:SoTA_comparison} reports overall and per-category average success rates and makespan. Figure~\ref{fig:sucess_rates_histogram} shows per-prompt success rates. \ourmethod achieved the highest overall success rate of $84.4\%$ with CoNVOI second with $50.0\%$. \ourmethod's strong performance on \textit{Coverage} tasks likely reflects its dedicated coverage planner, highlighting the benefit of adaptive planner selection. \ourmethod also outperforms the baselines on the other two task types, indicating a superior VLM prompting format. Successful paths produced by \ourmethod were on average 27\% shorter in makespan than CoNVOI marking (1.00 vs.\ 1.37), largely because \ourmethod assigns each subtask to the best-suited robot and divides large coverage tasks between the two. In contrast, the baselines frequently sent a robot to a goal closer to its teammate or assigned the same goals to both.

\begin{table}[h]
\centering
\caption{Comparison with baselines, averaged by task category.}
\label{tab:SoTA_comparison}
\begin{tabular}{lcccc}
\hline
\textbf{Method} & \textbf{Nav2Point} & \textbf{Coverage} & \textbf{Maneuver} & \textbf{All} \\
\hline
\multicolumn{5}{l}{\textit{Average Success Rate (\%) $\uparrow$}} \\
\hline
\ourmethod       & \textbf{85.0} & \textbf{80.0} & \textbf{88.3} & \textbf{84.4} \\
Pixel Selection \cite{bao2026goal2pixel,hu2025see}  & 26.7 & 25.0 & 43.3 & 31.7 \\
Grid Overlay \cite{pasqualina2026vision}     & 40.0 & 25.0 & 51.7 & 38.9 \\
CoNVOI Marking \cite{sathyamoorthy2024convoi}     & 45.0 & 45.0 & 60.0 & 50.0 \\
\hline
\multicolumn{5}{l}{\textit{Average Normalized Makespan $\downarrow$}} \\
\hline
\ourmethod       & \textbf{1.00} & \textbf{1.00} & 1.00 & \textbf{1.00} \\
Pixel Selection \cite{bao2026goal2pixel,hu2025see}  & 1.34 & 1.33 & \textbf{0.96} & 1.22 \\
Grid Overlay \cite{pasqualina2026vision}    & 1.58 & 1.85 & 1.44 & 1.61 \\
CoNVOI Marking \cite{sathyamoorthy2024convoi}   & 1.17 & 1.69 & 1.17 & 1.37 \\
\hline
\end{tabular}
\end{table}
\vspace{-12pt}

\subsection{Ablation Testing}\label{sec:ablation_testing}

Three different configurations of \ourmethod isolated the contribution of each component. \textit{CoPlanVLM W/O Obs. Marking} does not mark detected obstacles with the red ``x'' symbol as described in Section~\ref{sec:setofmarks} and the reasoning step where the VLM names the objects marked by each red ``x'' is accordingly skipped. \textit{CoPlanVLM Maneuver Only} retains the same marking, but does away with the task categorization and instead only uses the Maneuver planner which was chosen as the single planner best capable of completing all tasks since the Nav2Point planner is only used to navigate to a few set positions, and the Coverage planner cannot specify an order for the points to be visited. Finally, \textit{CoPlanVLM No CoT Reasoning} retains the original marking scheme and task categorizer, but removes the CoT reasoning trace, instead prompting the VLM to directly output waypoints.

The ablation study (Table~\ref{tab:ablation}) indicates that each of the three components contribute significantly to the overall success rate. Removing the marked obstacles reduced the Nav2Point and Maneuver success rates by $-40\%$ and $-18.3\%$ respectively, largely due to the VLM selecting inaccessible regions or missing crucial objects in the scene. Performance on Coverage tasks improved slightly ($+15\%$) without obstacle marking. We hypothesize that because the coverage planner drops any unsafe points, the VLM can safely select whole regions without reasoning about individual obstacles, so the additional obstacle-labeling step may introduce noise rather than useful information for these tasks. This supports extending the task specific prompting to adaptively select the best image overlay for each task. Using only the Maneuver prompt and planner for all tasks resulted in $-13.3\%$ SR and $-55\%$ SR on Nav2Point and Coverage tasks respectively along with a higher makespan indicating that relying on a single planner for all tasks reduces solution quality. Removing the CoT reasoning had the greatest overall effect with an average of $-35.5\%$ SR over all tasks.

\begin{table}[htb]
\centering
\setlength{\tabcolsep}{2pt}
\caption{Ablation study of \ourmethod, averaged by task category.}
\label{tab:ablation}
\begin{tabular}{lcccc}
\hline
\textbf{Method} & \textbf{Nav2Point} & \textbf{Coverage} & \textbf{Maneuver} & \textbf{All} \\
\hline
\multicolumn{5}{l}{\textit{Average Success Rate (\%) $\uparrow$}} \\
\hline
\ourmethod                 & \textbf{85.0} & 80.0 & \textbf{88.3} & \textbf{84.4} \\
\ourmethod w/o Marked Obs. & 45.0 & \textbf{95.0} & 70.0 & 70.0 \\
\ourmethod Maneuver Only   & 71.7 & 25.0 & \textbf{88.3} & 61.7 \\
\ourmethod w/o CoT         & 53.3 & 71.7 & 21.7 & 48.9 \\
\hline
\multicolumn{5}{l}{\textit{Average Normalized Makespan $\downarrow$}} \\
\hline
\ourmethod                 & \textbf{1.00} & \textbf{1.00} & 1.00 & \textbf{1.00} \\
\ourmethod w/o Marked Obs. & 1.43 & 1.10 & \textbf{0.99} & 1.15 \\
\ourmethod Maneuver Only   & 1.30 & 1.49 & 1.00 & 1.22 \\
\ourmethod w/o CoT         & 1.03 & 1.25 & 1.19 & 1.17 \\
\hline
\end{tabular}
\end{table}
\vspace{-10pt}

\begin{table}[htb]
\centering
\setlength{\tabcolsep}{4pt}
\caption{Hardware validation results.}
\label{tab:hardware_results}
\begin{tabular}{c l c c}
\hline
\textbf{Prompt} & \textbf{Category} &
\makecell{\textbf{Offline: Successful}\\\textbf{Plans (out of 10)}} &
\makecell{\textbf{Hardware: Successful}\\\textbf{Executions (out of 2)}} \\
\hline
4 & Nav2Point & 10 & 2 \\
5 & Nav2Point & 10 & 2 \\
6 & Nav2Point & 5  & 1 \\
10 & Coverage  & 9  & 2 \\
11 & Coverage  & 7  & 1 \\
12 & Coverage  & 7  & 2 \\
16 & Maneuver  & 10 & 2 \\
17 & Maneuver  & 10 & 2 \\
18 & Maneuver  & 5  & 1 \\
\hline
\multicolumn{2}{l}{\makecell[l]{\textbf{Avg.} \textbf{Success Rate (\%)}}} & \textbf{81.1} & \textbf{83.3} \\
\hline
\end{tabular}
\end{table}
\vspace{-10pt}

\subsection{Robot Deployment}\label{sec:robot_deployment}
\ourmethod was deployed on two Turtlebot 4 robots in an indoor robotics laboratory equipped with a motion capture room for robot pose estimation and an overhead camera. GPT-4o was used as both the classifier LLM and the executive VLM. A lab computer was used to perform all calculations, perform API calls to the VLM, and communicate linear and angular velocity commands to the Turtlebots through ROS2 Humble. All nine prompts tested offline on real-world images from the AVL were evaluated twice on hardware with fresh overhead images and newly generated plans for each trial. Hardware deployments (Table~\ref{tab:hardware_results}) demonstrated similar success rates to offline testing: $83.3\%$ vs. $81.1\%$ indicating that \ourmethod transfers smoothly to physical hardware. The three recorded failures during testing resulted either from missed obstacle detections (prompts 6 and 18) or incomplete coverage (prompt 11) and were visible in the plan before execution. Figure~\ref{fig:full_results_image} (Bottom) illustrates the paths planned and actual paths for three tasks.

\section{Conclusions, Limitations, and Future Work}

\ourmethod is a method for leveraging VLMs to perform long-horizon path planning for a collaborative team of two robots. \ourmethod improves the spatial reasoning, coordination, and path-planning capabilities of a base VLM through image preprocessing, task categorization, and structured chain-of-thought reasoning leading to improved success rates and more efficient paths compared to baselines.

Several limitations remain. Although \ourmethod's 84.4\% success rate outperforms existing approaches, reliability remains insufficient for mission-critical deployment. One common failure mode arises when CLIPSeg fails to detect small obstacles: an undetected box drove the 50\% success rate on prompts 6 and 18, suggesting \ourmethod could benefit from complementary sensing (e.g., LiDAR) to refine the occupancy map. \ourmethod also assumes a top-down view of the workspace, access to a cloud-hosted VLM, and was evaluated on only two robots. Future work will address these limitations by fine-tuning VLMs for waypoint selection, investigating edge deployment, and scaling to larger fleets.

\section*{ACKNOWLEDGMENT}
Generative AI tools (ChatGPT and Claude) were used to draft and debug portions of the code of \ourmethod. All code was validated by the authors.

\bibliographystyle{IEEEtran}
\bibliography{ref}

\end{document}

%% file: include/commands.tex
\newcommand{\ourmethod}{\textit{CoPlanVLM}\xspace}

\definecolor{darkgreen}{RGB}{0,120,0}

%% file: ref.bib
@inproceedings{koenig2004design,
  title={Design and use paradigms for gazebo, an open-source multi-robot simulator},
  author={Koenig, Nathan and Howard, Andrew},
  booktitle={2004 IEEE/RSJ international conference on intelligent robots and systems (IROS)(IEEE Cat. No. 04CH37566)},
  volume={3},
  pages={2149--2154},
  year={2004},
  organization={Ieee}
}

@article{bao2026goal2pixel,
  title={Goal2Pixel: Grounding Goals to Pixels for Vision-Language Navigation},
  author={Bao, Muyi and Cai, Yuxin and Xu, Hang and Li, Zongtai and He, Jinxi and Tang, Jingfan and Lv, Chen and Zhang, Ji and Xie, Yaqi and Wang, Wenshan},
  journal={arXiv preprint arXiv:2606.01621},
  year={2026}
}

@inproceedings{hu2025see,
  title={See, point, fly: A learning-free vlm framework for universal unmanned aerial navigation},
  author={Hu, Chih Yao and Lin, Yang-Sen and Lee, Yuna and Su, Chih-Hai and Lee, Jie-Ying and Tsai, Shr-Ruei and Lin, Chin-Yang and Chen, Kuan-Wen and Ke, Tsung-Wei and Liu, Yu-Lun},
  booktitle={Conference on Robot Learning},
  pages={4697--4708},
  year={2025},
  organization={PMLR}
}

@inproceedings{pasqualina2026vision,
  title={Vision-Language Model Informed Relative Navigation for GPS-Denied Environments},
  author={Pasqualina, Alexander and Peckham, Alison and Craddock, Caleb and Laxminarayan, Sneha and Winkleblech, Katelyn and Kumar, Sandip Sharan Senthil and Chhaya, Harshang and Paquet, Jared and Ramos, J Humberto and Anderson, Michael},
  booktitle={2026 IEEE Aerospace Conference},
  pages={1--11},
  year={2026},
  organization={IEEE}
}

@inproceedings{sathyamoorthy2024convoi,
  title={Convoi: Context-aware navigation using vision language models in outdoor and indoor environments},
  author={Sathyamoorthy, Adarsh Jagan and Weerakoon, Kasun and Elnoor, Mohamed and Zore, Anuj and Ichter, Brian and Xia, Fei and Tan, Jie and Yu, Wenhao and Manocha, Dinesh},
  booktitle={2024 IEEE/RSJ International Conference on Intelligent Robots and Systems (IROS)},
  pages={13837--13844},
  year={2024},
  organization={IEEE}
}

@inproceedings{luddecke2022image,
  title={Image segmentation using text and image prompts},
  author={L{\"u}ddecke, Timo and Ecker, Alexander},
  booktitle={2022 IEEE/CVF Conference on Computer Vision and Pattern Recognition (CVPR)},
  pages={7076--7086},
  year={2022},
  organization={IEEE}
}

@ARTICLE{astar,
  author={Hart, Peter E. and Nilsson, Nils J. and Raphael, Bertram},
  journal={IEEE Transactions on Systems Science and Cybernetics}, 
  title={A Formal Basis for the Heuristic Determination of Minimum Cost Paths}, 
  year={1968},
  volume={4},
  number={2},
  pages={100-107},
  doi={10.1109/TSSC.1968.300136}}

@article{rosenkrantz1977analysis,
  title={An analysis of several heuristics for the traveling salesman problem},
  author={Rosenkrantz, Daniel J and Stearns, Richard E and Lewis, II, Philip M},
  journal={SIAM journal on computing},
  volume={6},
  number={3},
  pages={563--581},
  year={1977},
  publisher={SIAM}
}

@article{croes1958method,
  title={A method for solving traveling-salesman problems},
  author={Croes, Georges A},
  journal={Operations research},
  volume={6},
  number={6},
  pages={791--812},
  year={1958},
  publisher={INFORMS}
}

@incollection{dijkstra2022note,
  title={A note on two problems in connexion with graphs},
  booktitle={Edsger Wybe Dijkstra: his life, work, and legacy},
  pages={287--290},
  year={2022}
}

@article{wilson2020robotarium,
  title={The robotarium: Globally impactful opportunities, challenges, and lessons learned in remote-access, distributed control of multirobot systems},
  author={Wilson, Sean and Glotfelter, Paul and Wang, Li and Mayya, Siddharth and Notomista, Gennaro and Mote, Mark and Egerstedt, Magnus},
  journal={IEEE Control Systems Magazine},
  volume={40},
  number={1},
  pages={26--44},
  year={2020},
  publisher={IEEE}
}

@article{borate2025CoMuRoS,
  title={LLM-based generalizable hierarchical task planning and execution for heterogeneous robot teams with event-driven replanning},
  author={Borate, Suraj and Pardeshi, Vipul and Vadali, Madhu and others},
  journal={arXiv preprint arXiv:2511.22354},
  year={2025}
}

@article{li2026survey,
  title={Large language models for multi-robot systems: A survey},
  author={Li, Peihan and An, Zijian and Abrar, Shams and Zhou, Lifeng},
  journal={Autonomous Robots},
  volume={50},
  number={3},
  pages={30},
  year={2026},
  publisher={Springer}
}

@inproceedings{liu2025coherent,
  title={Coherent: Collaboration of heterogeneous multi-robot system with large language models},
  author={Liu, Kehui and Tang, Zixin and Wang, Dong and Wang, Zhigang and Li, Xuelong and Zhao, Bin},
  booktitle={2025 IEEE International Conference on Robotics and Automation (ICRA)},
  pages={10208--10214},
  year={2025},
  organization={IEEE}
}

@INPROCEEDINGS{safety_aware_planning,
  author={Khan, Azal Ahmad and Andrev, Michael and Murtaza, Muhammad Ali and Aguilera, Sergio and Zhang, Rui and Ding, Jie and Hutchinson, Seth and Anwar, Ali},
  booktitle={2025 IEEE/RSJ International Conference on Intelligent Robots and Systems (IROS)}, 
  title={Safety Aware Task Planning via Large Language Models in Robotics}, 
  year={2025},
  volume={},
  number={},
  pages={21024-21031},
  doi={10.1109/IROS60139.2025.11246041}}

@article{yu2023conavgpt,
  title={Co-navgpt: Multi-robot cooperative visual semantic navigation using large language models},
  author={Yu, Bangguo and Kasaei, Hamidreza and Cao, Ming},
  journal={arXiv preprint arXiv:2310.07937},
  volume={3},
  year={2023}
}

@article{wang2025comres,
  title={COMRES-VLM: Coordinated Multi-Robot Exploration and Search using Vision Language Models},
  author={Wang, Ruiyang and Hsu, Hao-Lun and Hunt, David and Kim, Jiwoo and Luo, Shaocheng and Pajic, Miroslav},
  journal={arXiv preprint arXiv:2509.26324},
  year={2025}
}

@inproceedings{anderson2018vln,
  title={Vision-and-language navigation: Interpreting visually-grounded navigation instructions in real environments},
  author={Anderson, Peter and Wu, Qi and Teney, Damien and Bruce, Jake and Johnson, Mark and S{\"u}nderhauf, Niko and Reid, Ian and Gould, Stephen and Van Den Hengel, Anton},
  booktitle={2018 IEEE/CVF conference on computer vision and pattern recognition},
  pages={3674--3683},
  year={2018},
  organization={IEEE}
}

@article{goetting2024end,
  title={End-to-end navigation with vision language models: Transforming spatial reasoning into question-answering},
  author={Goetting, Dylan and Singh, Himanshu Gaurav and Loquercio, Antonio},
  journal={arXiv preprint arXiv:2411.05755},
  year={2024}
}

@article{tian2024drivevlm,
  title={Drivevlm: The convergence of autonomous driving and large vision-language models},
  author={Tian, Xiaoyu and Gu, Junru and Li, Bailin and Liu, Yicheng and Wang, Yang and Zhao, Zhiyong and Zhan, Kun and Jia, Peng and Lang, Xianpeng and Zhao, Hang},
  journal={arXiv preprint arXiv:2402.12289},
  year={2024}
}

@article{cheng2024navila,
  title={Navila: Legged robot vision-language-action model for navigation},
  author={Cheng, An-Chieh and Ji, Yandong and Yang, Zhaojing and Gongye, Zaitian and Zou, Xueyan and Kautz, Jan and B{\i}y{\i}k, Erdem and Yin, Hongxu and Liu, Sifei and Wang, Xiaolong},
  journal={arXiv preprint arXiv:2412.04453},
  year={2024}
}

@article{wang2025trackvla,
  title={Trackvla: Embodied visual tracking in the wild},
  author={Wang, Shaoan and Zhang, Jiazhao and Li, Minghan and Liu, Jiahang and Li, Anqi and Wu, Kui and Zhong, Fangwei and Yu, Junzhi and Zhang, Zhizheng and Wang, He},
  journal={arXiv preprint arXiv:2505.23189},
  year={2025}
}

@article{rana2026overseec,
  title={OVerSeeC: Open-Vocabulary Costmap Generation from Satellite Images and Natural Language},
  author={Rana, Rwik and Quattrociocchi, Jesse and Lee, Dongmyeong and Ellis, Christian and Adkins, Amanda and Uccello, Adam and Warnell, Garrett and Biswas, Joydeep},
  journal={arXiv preprint arXiv:2602.18606},
  year={2026}
}

@article{potnis2026catnav,
  title={CATNAV: Cached Vision-Language Traversability for Efficient Zero-Shot Robot Navigation},
  author={Potnis, Aditya and Affonso, Francisco and Gummadi, Shreya and Uppalapati, Naveen Kumar and Chowdhary, Girish},
  journal={arXiv preprint arXiv:2603.22800},
  year={2026}
}

@inproceedings{elnoor2025gronav,
  title={Vlm-gronav: Robot navigation using physically grounded vision-language models in outdoor environments},
  author={Elnoor, Mohamed and Weerakoon, Kasun and Seneviratne, Gershom and Xian, Ruiqi and Guan, Tianrui and Jaffar, Mohamed Khalid M and Rajagopal, Vignesh and Manocha, Dinesh},
  booktitle={2025 IEEE International Conference on Robotics and Automation (ICRA)},
  pages={2391--2398},
  year={2025},
  organization={IEEE}
}

@article{stogiannidis2025mind,
  title={Mind the gap: Benchmarking spatial reasoning in vision-language models},
  author={Stogiannidis, Ilias and McDonagh, Steven and Tsaftaris, Sotirios A},
  journal={arXiv preprint arXiv:2503.19707},
  year={2025}
}

@article{yang2023setofmark,
  title={Set-of-mark prompting unleashes extraordinary visual grounding in gpt-4v},
  author={Yang, Jianwei and Zhang, Hao and Li, Feng and Zou, Xueyan and Li, Chunyuan and Gao, Jianfeng},
  journal={arXiv preprint arXiv:2310.11441},
  year={2023}
}

@inproceedings{bai2025room2chess,
  title={R2C: Mapping Room to Chessboard to Unlock LLM As Low-Level Action Planner},
  author={Bai, Ziyi and Li, Hanxuan and Fu, Bin and Xiong, Chuyan and Wang, Ruiping and Chen, Xilin},
  booktitle={2025 IEEE/CVF Conference on Computer Vision and Pattern Recognition (CVPR)},
  pages={19456--19466},
  year={2025},
  organization={IEEE}
}

@article{agaskar2025deepfleet,
  title={DeepFleet: Multi-agent foundation models for mobile robots},
  author={Agaskar, Ameya and Siva, Sriram and Pickering, William and O'Brien, Kyle and Kekeh, Charles and Barbosa, Alexandre Ormiga Galvao and Li, Ang and Sarker, Brianna Gallo and Chua, Alicia and Nemade, Mayur and others},
  journal={arXiv preprint arXiv:2508.08574},
  year={2025}
}

@article{rahman2025multiuavsurvey,
  title={A survey on multi-UAV path planning: Classification, algorithms, open research problems, and future directions},
  author={Rahman, Mamunur and Sarkar, Nurul I and Lutui, Raymond},
  journal={Drones},
  volume={9},
  number={4},
  pages={263},
  year={2025},
  publisher={MDPI}
}

@article{karam2026collaboration,
  title={Collaboration in multi-robot systems: Taxonomy and survey over frameworks for collaboration},
  author={Karam, Riwa and Nguyen, Alexander A and Lin, Ruoyu and Martin, David R and Morales, Diana and Butler, Brooks A and Egerstedt, Magnus},
  journal={arXiv preprint arXiv:2603.23898},
  year={2026}
}
